\documentclass[sn-mathphys,Numbered]{sn-jnl}% Math and Physical Sciences Reference Style
\usepackage{graphicx}%
\usepackage{multirow}%
\usepackage{amsmath,amssymb,amsfonts}%
\usepackage{amsthm}%
\usepackage{mathrsfs}%
\usepackage[title]{appendix}%
\usepackage{xcolor}%
\usepackage{textcomp}%
\usepackage{manyfoot}%
\usepackage{booktabs}%
\usepackage{algorithm}%
\usepackage{algorithmicx}%
\usepackage{algpseudocode}%
\usepackage{listings}%
\usepackage{caption}
\usepackage{subcaption}
\usepackage{tabularx}
\usepackage{changepage}

\theoremstyle{thmstyleone}%
\theoremstyle{thmstyletwo}%

\theoremstyle{thmstylethree}%

\begin{document}

% \title{{Analysing XAI applied to clinical prediction models: How much can I trust?}}
% \title{{Analysing XAI applied to clinical prediction models: Concordance Between Methods and Agreement with Future Events}}
% \title{{Analysing XAI applied to clinical prediction models: Untrustworthy but could be useful}}
% \title{{Consistency analysis and benchmarking of XAI applied to clinical prediction models}}

\title{{Evaluation of Popular XAI Applied to Clinical Prediction Models: Can They be Trusted?}}

%%=============================================================%%
%% Prefix	-> \pfx{Dr}
%% GivenName	-> \fnm{Joergen W.}
%% Particle	-> \spfx{van der} -> surname prefix
%% FamilyName	-> \sur{Ploeg}
%% Suffix	-> \sfx{IV}
%% NatureName	-> \tanm{Poet Laureate} -> Title after name
%% Degrees	-> \dgr{MSc, PhD}
%% \author*[1,2]{\pfx{Dr} \fnm{Joergen W.} \spfx{van der} \sur{Ploeg} \sfx{IV} \tanm{Poet Laureate} 
%%                 \dgr{MSc, PhD}}\email{iauthor@gmail.com}
%%=============================================================%%

\author*[1]{\fnm{Aida} \sur{Brankovic}}\email{aida.brankovic@csiro.au}

\author[2]{\fnm{David} \sur{Cook}}\email{david.cook@health.qld.gov.au}
% % \equalcont{These authors contributed equally to this work.}

\author[1]{\fnm{Jessica} \sur{Rahman}}\email{jessica.rahman@csiro.au}
% % \equalcont{These authors contributed equally to this work.}
\author[3]{\fnm{Wenjie} \sur{Huang}} \email{wenjie.huang@uq.net.au}

\author[1]{\fnm{Sankalp} \sur{Khanna}} \email{sankalp.khanna@csiro.au}

% \affil*[1]{\orgdiv{Department}, \orgname{Organization}, \orgaddress{\street{Street}, \city{City}, \postcode{100190}, \state{State}, \country{Country}}}

% \affil[2]{\orgdiv{Department}, \orgname{Organization}, \orgaddress{\street{Street}, \city{City}, \postcode{10587}, \state{State}, \country{Country}}}

% \affil[3]{\orgdiv{Department}, \orgname{Organization}, \orgaddress{\street{Street}, \city{City}, \postcode{610101}, \state{State}, \country{Country}}}

\affil[1]{CSIRO Australian e-Health Research Centre, Brisbane, QLD 4029, Australia}
\affil[2]{Intensive Care Unit, Princess Alexandra Hospital, Brisbane, QLD 4102, Australia}
\affil[3]{The University of Queensland}

%%==================================%%
%% sample for unstructured abstract %%
%%==================================%%

\abstract{The absence of transparency and explainability hinders the clinical adoption of Machine learning (ML) algorithms. Although various methods of explainable artificial intelligence (XAI) have been suggested, there is a lack of literature that delves into their practicality and assesses them based on criteria that could foster trust in clinical environments. 
%their usefulness and elements contributing to it such as explanation concordance, consistency, and agreement of generated explanations with expert clinical knowledge is sporadic. 
To address this gap this study evaluates two popular XAI methods used for explaining predictive models in the healthcare context in terms of whether they (i) generate domain-appropriate representation, i.e. coherent with respect to the application task, (ii) impact clinical workflow and (iii) are consistent. To that end, explanations generated at the cohort and patient levels were analysed. The paper reports the first benchmarking of the XAI methods applied to risk prediction models obtained by evaluating the concordance between generated explanations and the trigger of a future clinical deterioration episode recorded by the data collection system. We carried out an analysis using two Electronic Medical Records (EMR) datasets sourced from Australian major hospitals. The findings underscore the limitations of state-of-the-art XAI methods in the clinical context  and their potential benefits. We discuss these limitations and contribute to the theoretical development of trustworthy XAI solutions where clinical decision support guides the choice of intervention by suggesting the pattern or drivers for clinical deterioration in the future.}


%%\pacs[JEL Classification]{D8, H51}

%%\pacs[MSC Classification]{35A01, 65L10, 65L12, 65L20, 65L70}

\maketitle

\section{Introduction}\label{sec1}
ML-based tools have the potential to significantly improve health and healthcare delivery  \cite{callahan2017machine}, yet these methods are often "black-box" in nature. In this context, ML models often fail to elucidate which influential factors affect individual predictions as well as how changes to these observable inputs affect or modulate the outcome being predicted. This is an important deficiency because clinicians' understanding and  confidence  in using predictions to guide interventions in a complex process that is modelled is the key to trust. Knowing the main contributing factors allows their evaluation in terms of their coherence with respect to the application task and its potential actionability could help in building trust in clinical settings \cite{tonekaboni2019clinicians}. If the explanation cannot explain why a future problem will evolve, and thus justify treatment interventions then clinicians can rightly be sceptical. The lack of transparency in their function reduces their trustworthiness \cite{tjoa2020survey, vayena2018machine} and is a barrier to their adoption for clinical decision-making. Understanding of the methods should also be sufficient for clinician users to suspect when the tools are not working, or being used outside the purpose for which they were developed. 

To understand complex ML models, several eXplainable AI (XAI) methods have been proposed in the literature \cite{tjoa2020survey}. These methods can be categorised into (i) \textit{gradient-based} e.g., SmoothGrad \cite{smilkov2017smoothgrad}, Integrated Gradients \cite{sundararajan2017axiomatic}, Deep Taylor Decomposition (DTD) \cite{montavon2017explaining}), Layer-wise propagation \cite{montavon2019layer}, and (ii) \textit{perturbation-based} e.g. LIME \cite{ribeiro2016should}, Shap \cite{lundberg2020local}. However, there is little understanding of how applicable or useful they are in clinical settings or whether they should be significantly re-tailored or even novel XAI methods developed.

From clinicians’ view, knowing the subset of features driving the model outputs is crucial as it allows them to compare the data-driven model decisions to their clinical judgment, especially important in case of a disagreement \cite{tonekaboni2019clinicians}. Tonekaboni et. al. \cite{tonekaboni2019clinicians} also suggest that rigorous evaluation of explanations against the following criteria: (i) domain-appropriate representation, (ii) potential actionability and (iii) consistency could contribute to building trust in clinical settings. Other recent studies \cite{krishna2022disagreement, brankovic2023medinfo} report inconsistency between the explanations generated by various popular explanation methods. This variability implies that at least some generated explanations are incorrect. If incorrect, explanations at a patient level could be misleading and could lead to wrong decisions with dire consequences in applications such as healthcare. Thus, to build trust, it is critical to investigate the conformity of explanations with clinical expert knowledge by evaluating them against the aforementioned criteria.

This paper presents the results of the quantitative analysis of explanations at a patient (i.e. local explanations) and cohort (i.e. global explanations) level and discusses them in terms of their coherence with respect to the application task, impact on the workflow and consistency. To investigate the utility and the trustworthiness of the XAI-generated explanations in the clinical context, we evaluate against criteria suggested by Tonekaboni et al. \cite{tonekaboni2019clinicians}. To analyse discordance between explanations, we employ agreement metrics proposed by Krishna et. al. \cite{krishna2022disagreement} where appropriate. The analysis is performed on two EMR datasets sourced from two major Australian hospitals examining data-driven models predicting unexpected patient deterioration \cite{brankovic2022explainable} and hospital readmission after discharge \cite{brankovic2022identifying}.  

We used Shap and DTD methods to generate explanations of tree-based and Neural Network (NN)-based ML models and patient and cohort level. These explanations were compared to each other and also to interpretations arising from the coefficients of the logistic regression (LR) model, the most accepted predictive model in healthcare applications. The explanations obtained for one of the two datasets used in this study (for which it was possible) were also benchmarked against the true causes recorded by the deployed data collection system in the study hospital. We discuss these results and their implications from clinicians’ perspectives. The necessary criteria for having trustworthy explanations and how these guide the choice of intervention are also considered.

\section{Related work}
This work builds on (i) XAI methods introduced in the literature, and (ii) work related to exploring the properties of the existing methods.  
Considering targeted applications, here we focus on commonly deployed XAI for predictive models in healthcare.  Models such as Logistic regression and rule-based (e.g. decision trees) allow insight into a model structure and support the interpretation of the decision they make. However, there is evidence \cite{sufriyana2020comparison, shin2021machine} that more complex models such as random forest, gradient-boosted trees (XGB) or deep neural networks (DNN) might perform better albeit at the cost of being ad hoc nontransparent. Consequently, a number of methods have been proposed for generating post hoc explanations for these "black-box" models. Shap \cite{lundberg2020local} is the most popular method used for explaining tree-based predictive models in healthcare \cite{brankovic2022explainable, brankovic2022identifying},\cite{cummings2021predicting, lejarza2021optimal,moreno2020development, zhou2022exploring, wang2021interpretable, song2020cross}. The method explains the prediction by computing the contribution of each feature to the model’s prediction using game theory as a theoretical background. As such can be used to explain prediction obtained for a single patient, or if summed up across all patients it provides global feature importance. Lauritsen et al. \cite{lauritsen2020explainable} deployed DTD concept \cite{montavon2017explaining}, originally developed for image classification problems to explain DNN model developed to predict acute critical illness. The DTD method explains the prediction by applying Taylor decomposition per each neuron.

Krishna et al. \cite{krishna2022disagreement} investigated the agreement between the explanations generated by different popular explanation methods and reported low concordance. The authors recommend a systematic study of the reasons behind the occurrence of the explanation disagreement problem. 
Literature investigating the discordance, usability and reliability of these methods on models developed using clinical data is extremely scarce. Disagreement and the reasons behind it in the  context of clinical predictive models are investigated and discussed in \cite{brankovic2023medinfo}. Saporta et al. \cite{saporta2022benchmarking} provided the first human benchmarking methods based on saliency maps for human chest X-ray interpretation. Clinicians' expectations of explanations and the understanding of when explainability helps to improve clinicians' trust are studied in \cite{tonekaboni2019clinicians}.  Some limitations of explainability methods from the cardiologist's perspective were discussed in \cite{petch2022opening}. Authors in \cite{ghanvatkar2022towards} and \cite{tonekaboni2019clinicians} attempted to address the development of criteria for the evaluation of the explanations in terms of their usefulness to clinicians.  However, none of them concentrates on evaluating explanations, their usability and their effect on trust in clinical settings.

\section{Methods}
\subsection{Datasets}
This study was conducted under institutional ethics approval (Metro South Human Research Ethics Committee HREC/18/QPAH/525 and Ref HREC/16/QPAH/217). Analysis was performed in accordance with the guidelines and regulations on deidentified data. The previous studies \cite{brankovic2022explainable, brankovic2022identifying} focused on two predictive algorithms - one for predicting patient deterioration events in an acute adult inpatient ward setting and the other for predicting patient readmission following discharge in an inpatient pediatric setting.

\subsection{Data used in analysis, predictors and response variables}

The dataset used for predicting patient deterioration, hereafter referred to as the Vital signs (VS) dataset \cite{brankovic2022explainable}, was restricted to adult inpatient ward patients with a detailed cohort selection procedure reported in \cite{brankovic2022explainable}. In this study, we used de-identified data from 1/1/2016 to 31/1/2018 for modelling and explainer development \cite{brankovic2022explainable}. Vital signs (Systolic and Diastolic Blood Pressure, Mean Arterial Pressure, Heart Rate, Temperature, Respiratory Rate, Oxygen Saturation (SpO2), and Oxygen flow rate (O2 flow rate), level of consciousness (AVPU), and their derivatives, demographic information (age, gender and the patient’s length of stay (LOS)) were used as predictors. Red flag deterioration alert in the next 8 hours represented the predicted outcome of interest. 

The dataset used to predict readmission within the next 30 days, hereafter referred to as the RA30 dataset \cite{brankovic2022identifying} is sourced de-identified inpatient pediatric administrative data from 1/1/2019 to 31/12/2020 for modelling and explainer development. Predictors used in this study were demographics, patient stay history as well as medication-related and pathology-related variables. Readmission within the next 30 days (RA30) was the predicted outcome.

For details related to both datasets the reader is referred to \cite{brankovic2022explainable, brankovic2022identifying}.

\subsection{Predictive models}
To explore the agreement between the features obtained by different methods we selected three modelling approaches, each as a representative of one modelling paradigm: regression, conventional ML, and deep neural network (DNN). We chose logistic regression (LR) with l1 regularization (L1), aka LASSO, as the most accepted predictive model for clinical applications. Unlike other ML models which are generally more complex, it allows insight into the model structure and is a trusted and accepted model among clinicians. We considered XGB as representative of conventional ML which along with the random forest demonstrated to be powerful models when dealing with large and complex data. We deployed a dense NN model to explore the explainability of NN-based models.

\subsection{Explanations}
Shap and DTD explainers are constructed on top of the models developed using the VS and RA30 datasets. To measure agreement, we compared generated explanations at the patient and cohort levels. Patient-wise explanations (i.e.  local explanations) are those computed for each patient individually for each predicted outcome. To gain an insight into features that underpin decisions of a predictive model overall, we analysed explanations at the cohort level (i.e. global explanations) which are obtained by adding up the absolute values of all explanations obtained for individual patients and averaging them over the total number of considered patients. \\%Considering that the focus of risk predictive models is on identifying the patients at risk of deterioration or readmission we considered only correctly predicted patients for computing the \textit{global} and \textit{local} feature relevance.  \\

\subsection{ Agreement Quantification }
Metrics for measuring the agreement between the explanations introduced in \cite{krishna2022disagreement} were used in different experiments carried on to evaluate explanations intended to build trust in clinical settings. This subsection solely defines agreement metrics that were employed in the analysis, while the successive subsection explains and justifies their utilization.

An intuitive measure of the agreement between the explanations obtained with different methods is the number of common features in sets of top features identified by different methods which will be denoted in the remainder of the paper as the \textit{feature agreement (FA)}. To quantify the agreement in the direction features contribute to the prediction we compared explanation signs when performing a patient-wise analysis. However, \textit{sign} analysis was considered only in cases where the explanations can take positive and negative values. Though \textit{Ranking Agreement (RA)} was proposed as one of the metrics for agreement quantification \cite{krishna2022disagreement} it has not been included. Considering that there is a degree of interdependence between the features, in particular pathophysiology represented in the vital signs, causing the variation in top features it was advised by the clinical collaborator as a non-informative metric. Instead, \textit{Ranking correlation (CR)} was recognised as more useful as it allows understanding the ranking agreement over the whole feature set where there are complex interdependencies. 

Considered agreement metrics are defined in Krishna et al. \cite{krishna2022disagreement} as follows:\\
 $\bullet$ \textbf{\textit{FA}} represents the fraction of common features between the sets of top-n features obtained by two explanation methods.\\
 $\bullet$ \textbf{\textit{CR}} is quantified by Spearman's rank correlation coefficient. It is used to assess the strength and direction of association between two ranked variables, i.e. feature ranking obtained by two different methods.\\
 $\bullet$ \textbf{\textit{sign}} quantifies the fraction of common features in the sets of top-n that have the same sign, which shows the direction of contribution. As such it is used only for patient-level analysis. Considering that DTD is positive, it has been only computed for the Shap.

\subsection{Experiment design}
 To evaluate explanations in the context of clinical predictive models against the criteria suggested in \cite{tonekaboni2019clinicians} we used explanations obtained for EMR-based predictive models reported in \cite{brankovic2022explainable}, \cite{brankovic2022identifying}. To that end extensive analysis is performed to investigate whether explanations (i) have domain-appropriate representation, i.e. whether their representation is coherent with respect to the application task, (ii) they may impact clinical workflow and (iii) are consistent. 
 
To examine whether the explanations are coherent with respect to the application task i.e. whether they are domain-appropriate, explanations at the cohort level (all samples considered) were taken into account. Explanations that are redundant are not desirable unless critical to potential clinical workflows, i.e. they should not further obfuscate model behaviour for the clinician \cite{tonekaboni2019clinicians}. To that end, top contributors at the cohort level obtained by different methods were compared and their representation in the context of the application task is assessed based on clinical expert knowledge. To account for the inherent randomness in the optimization procedure, analysis is performed on results obtained by running each algorithm five times. 

\begin{figure}[h!]
    \centering
    {\includegraphics[width=0.8\textwidth]{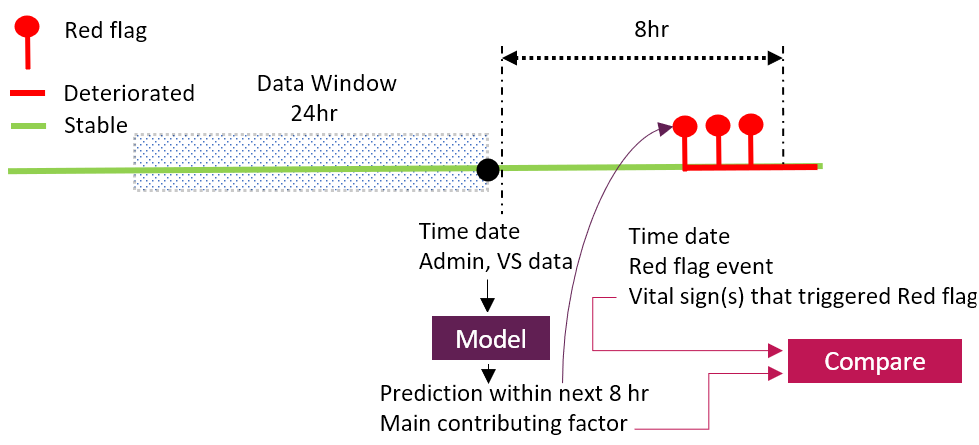}} 
    \caption{Visualisation of the benchmarking process.} 
    \label{fig:benchScheme}  
\end{figure}

To evaluate the potential actionability of the explanations i.e. whether they are informative and may impact the workflow by informing follow-up clinical workflow while at the same time being parsimonious and timely, explanations at the patient level were analysed, assessed and discussed. In this scenario, only the samples that correctly predicted patient deterioration/readmission were taken into consideration. Doing a patient-level analysis allowed the evaluation of explanations generated for each patient individually, assessment of their informativeness and their potential impact on the workflow. E.g. explanation of the main contributing factors can guide the choice of intervention and the risk assessment tool will help to allocate resources and explanations that should facilitate the decision-making \cite{tonekaboni2019clinicians}. 
Explanations intended to build trust in clinical settings could benefit from being rigorously evaluated against the ground truth. With that objective in mind, local explanations obtained for VS datasets were benchmarked against the red flag triggers recorded by the data collection system deployed in the PA hospital. We compared whether the top feature (either the original feature (e.g. SpO2) or its derivatives e.g. its min, max, std, average, count, slope) are matching the cause of the correctly predicted future red flag event recorded by the deployed system (one of the following measurements: blood pressure, pulse, oxygen saturation (SpO2), GCS, Sedation score or respiratory rate). In the analysis, only correct predictions were considered. Multiple red flags could potentially arise within the prediction window, but only the first one was taken into account (see Fig. \ref{fig:benchScheme}). Concordance was computed as a fraction of the total count of samples for which the explanation was matching the trigger recorded by the deployed data collection system and the total number of correctly predicted red flags. Given that none of the predictors in RA30 dataset offers a cause for patient readmission which could be considered as a ground truth, RA30 detest was excluded from benchmarking analysis. Considering that different methods may produce different results \cite{brankovic2023medinfo} and the discordance could impact clinicians' trust, explanation agreement between the methods at the patient has been also investigated and discussed.

Explanations should be consistent, i.e. they should (i) yield observable changes for any changes in predictions due to changes in inputs, (ii) be invariant to underlying design variations, i.e. they should only reflect relevant clinical variability. Violating any of these elements results in inconsistent explanations. This undermines their reliable actionability which in turn has a negative impact on clinicians' trust. \cite{tonekaboni2019clinicians} The first consistency factor suggests causality. However, though considered XAIs make correlations transparent these are patterns and not causal. On occasions, the patterns may be casual, but this is not to be expected. Kjersti et al. \cite{aas2021explaining} showed that marginal Shapley values may lead to incorrect explanations when features are highly correlated. While DTD can provide insights into how a neural network processes information by decomposing the NN into a sum of simple interpretable functions and associating them with a specific input feature to score each feature, it is not specifically designed for causal inference. 

To examine the consistency of explanations in relation to variations in the design of underlying models (DNN and XGB), explanations calculated at the cohort level were analyzed for each of the five independent runs. Obtained explanations were compared across the runs and their agreement was quantified with FA and CR metrics. 
Unlike DTD which is applicable only to NN, Shap explainer can be applied to any model. Therefore, the consistency of Shap explanations in relation to the deployed models, namely XGB and NN, was analysed and quantified with FA and CR metrics. \\ Ultimately, to explore the consistency of explanations regarding the direction of contribution (i.e., sign), explanations obtained at the patient level for both XGB and DNN models using the Shap method were analyzed across five independent runs.

\section{Results}

\subsection{Domain Appropriate Representation}
 Tab. \ref{tab:0} lists the top 5 features at the cohort level obtained after the first run of each algorithm. Rank lists for all 5 runs are provided in the Supplementary information Tab. S1-S10 and detailed reporting on their agreement in Supplementary Information Tab. S11-S18.

\begin{table}[h]
% \centering
\footnotesize
\begin{tabular}{cccc}
\hline
\textbf{VS}: Coeff L1     & \textbf{VS}: DNN Shap     & \textbf{VS}: DNN DTD      & \textbf{VS}: GB Shap     \\ \hline
LOS$^*$          & SpO2         & num. rec. VS & LOS             \\
SpO2$^*$         & LOS          & LOS          & SpO2          \\
Resp. Rate           & SBP $^*$          & SpO2         & SBP      \\
minSBP$^*$       & num. rec. VS & SBP          & num. rec. VS \\
DBP$^*$          & O2 flow rate        & O2 flow rate        & minSpO2      \\ \hline 

\textbf{RA30}: Coeff L1    & \textbf{RA30}: DNN Shap     & \textbf{RA30}: DNN DTD  & \textbf{RA30}: XGB Shap     \\ \hline
Prev. inpat. stay count              & Prev. inpat. stay count              & Prev. inpat. stay count              & Prev. inpat. stay count                 \\
Prev. inpat. stay count$^2$         & Patho.: No unique tests              & Patho.: No unique tests              & Patho.: No unique tests                 \\
Patho.: No unique tests              & Elec. status: NA        & Adm. source: Broader & Elec. status: NA$^*$              \\
Adm. source: Broader                 & LOS \textasciicircum{}2              & LOS \textasciicircum{}2              & LOS                                     \\
Elec. status: NA           & Elec. status: Emerg. admis.       & Adm. source: Emerg.dep.    & Path.: No.pat.observable codes       \\ \hline
&&& \\
\end{tabular} 
* LOS -Length of stay, SpO2- Oxygen saturation, SBP- Systolic Blood Pressure, DBP- Diastolic Blood Pressure, NA -Not Assigned 
\caption{Run 1: Top contributors generated by each of the considered methods for VS and RA30 datasets.} \label{tab:0}
\end{table}

\subsection{Impact on the Clinical Workflow}
\begin{figure}[h!]
    \centering
    {\includegraphics[width=0.38\textwidth]{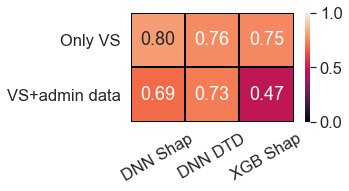}} 
    \caption{Fraction where explanations of  the most contributing predictors  match the red flag trigger events.} 
    \label{fig:8}  
\end{figure}

\textbf{Benchmarking}
The fraction of the samples for which the explanations and true flag triggers were matching is shown in Fig. \ref{fig:8}. The top row shows the fraction of samples that match the ground truth when correctly predicted samples that have an administrative feature (e.g. LOS) as the main contributor were excluded from comparison as they cannot be directly benchmarked against the causes recorded by the EWS. The percentage of excluded samples was 2.7\% for DNN-DTD, 13\% for DNN-Shap 13\% and 37\% for XGB-Shap.
Bottom row reports matching when samples with an administrative feature as the main predictors were considered. 

\textbf{Patient-wise concordance of top contributors}
% Table \ref{tab:7} lists the top 5 contributors for two randomly selected patients for who the algorithms correctly predicted deterioration within a single run. A list of all features sorted by importance is provided in Supplementary Information Tab. S23-S24. \\

% \begin{table}[]
% \centering
% \footnotesize
% \begin{tabular}{ccc|ccc}
% \hline
% \multicolumn{3}{c}{\textbf{Patient 1}} & \multicolumn{3}{c}{\textbf{Patient 2}} \\
% \hline
% DNN Shap           & DNN DTD      & XGB Shap         & DNN Shap         & DNN DTD         & XGB Shap        \\
% \hline
% Resp Rate          & Resp Rate    & Resp Rate        & lastvalue SpO2   & lastvalue SpO2  & lastvalue SpO2  \\
% O2 Flow Rate       & HeartRate    & LOS              & min SpO2         & lastvalue SBP   & min SpO2        \\
% mean Temp          & min Temp     & min HeartRate    & max SpO2         & 50\% SBP        & mean SpO2       \\
% mean Resp Rate     & mean Temp    & max Resp Rate    & O2 Flow Rate     & min SBP         & O2 Flow Rate    \\
% min HeartRate      & SBP          & std HeartRate    & mean Temp        & min Temp        & 50\% SpO2     \\
% \hline
% \end{tabular}

% \caption{VS: Top 5 contributors for two randomly selected deteriorating patients} \label{tab:7}

% \end{table}

The agreement on the top contributor across all methods over 5 independent runs for VS dataset was $84\% $ when we compared the exact names of the features (e.g. min SpO2) and $82\%$ when we compared whether the top features are derivatives of the same vital sign (e.g. SpO2). If the top two features are considered (the order did not matter), the incidence was $20\%$ and if the top 3 are considered it was less than $3\%$.  
For the readmission dataset, the agreement on the most contributing factor was $55\%$ and for the top two was less than $1\%$.\\
Detailed results of  agreement analysis are reported in Supplementary Information Tab. S25-26.

\subsection{Consistency}

\textbf{Sensitivity on underlying design variations} 
Average agreement metrics are reported in  Fig. \ref{fig:2}a and \ref{fig:2}b. For detailed results see Supplementary Information Tab. S18-S22.

\textbf{Sensitivity on deployed model}  Results shown in Fig. \ref{fig:2}c can be also used to gain insight into the consistency of the Shap method with respect to the model candidate. 

\textbf{Agreement on the direction of contribution}
For the RA30 dataset, the average \textit{sign} agreement for the top 5 contributors obtained with Shap on top of the DNN and XGB models over 5 different runs was 0.75 and 0.97 for the leading contributor. Agreement across the whole feature set was 0.41. For VS dataset, the \textit{sign} agreement for the major contributor was 0.99, for the top 5 0.86 and across the whole feature set was 0.58.

\begin{figure}[h!]
    \centering
     \begin{subfigure}[t]{0.26\textwidth}
         \centering
         \includegraphics[width=\textwidth]{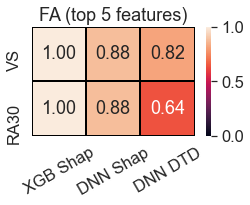}
         \caption{}
         % \label{fig:y equals x}
     \end{subfigure}
     % \hfill
     \begin{subfigure}[t]{0.26\textwidth}
         \centering
         \includegraphics[width=\textwidth]{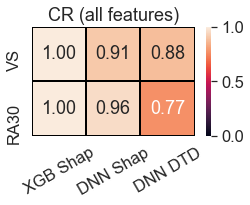}
         \caption{}
         % \label{tab:3} 
     \end{subfigure}
     % \hfill
     \begin{subfigure}[t]{0.22\textwidth}
         \centering
         \includegraphics[width=\textwidth]{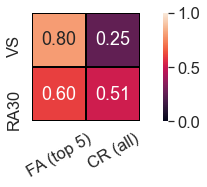}
         \caption{}
         % \label{tab:3} 
     \end{subfigure}
        % \caption{}
        % \label{tab:5}
    % {\includegraphics[width=0.29\textwidth]{FA_sameModelExplainer .png}} 
    % {\includegraphics[width=0.29\textwidth]{CR_sameModelExplainer .png}} 
    \caption{Consistency analysis: Sensitivity on underlying design variations represented with an average agreement of explanations obtained at the cohort level for the same model-explainer pairs across 5 independent runs (a, b). Shap sensitivity on the deployed model (c).} 
    \label{fig:2}
\end{figure}

\section{Discussion}
%Opening the Black Box: The Promise and Limitations of Explainable Machine Learning in Cardiology
Several studies \cite{tonekaboni2019clinicians,liu2022does,diprose2020physician,panigutti2022understanding} have explored the role of explainability in AI-based clinical decision-support tools. Liu et al. \cite{liu2022does} showed that physicians identified XAI to be very important in their perceived value and trust in the technology. Diprose et al. \cite{diprose2020physician} showed through responses from a hypothetical scenario that physicians trusted the outputs from two model-agnostic explainability methods over models with no explanation. Both studies assumed explanation consistency. However, explanation consistency cannot be assumed and a recent study \cite{krishna2022disagreement} showed that state-of-the-art explanation methods often disagree in terms of the explanations they output. The aim of this study was to assess the explanations generated by popular XAI methods used for explaining clinical predictive models. These explanations were evaluated against the metrics proposed in \cite{tonekaboni2019clinicians}, which are intended to contribute to the establishment of trust in clinical settings. For that purpose, 2 real-world datasets, 3 models and 2 popular XAI methods were deployed. In the following, we discuss the results in the context of the metrics employed i) domain-appropriate representation, (ii) potential actionability i.e. impact on a clinical workflow and (iii) consistency.\\
% Appropriate Representation 
% -Global explanations, top predictors 
In assessing domain-appropriate representation, predictions and related explanations were evaluated against domain expert knowledge. From a clinical perspective, influential predictors to anticipate patient deterioration aligned with vital signs parameters identified as important by clinical collaborators. These included observable vital signs such as AVPU (level of consciousness), respiratory function (SpO2, resp Rate and Oxygen therapy), cardiovascular function (systolic and diastolic BP); metadata of bedside activity indicating higher levels of existing care (e.g. the number of recorded vital signs and bedside events logged, and baseline measures of chronicity of the acute process (e.g. current inpatient length of stay). Similarly, for predicting patient readmission, informative predictors suggested by senior health care administrators including previous inpatient visit counts, previous emergency visit counts (stays and presentations), age, index of socioeconomic status and indicator for a stay longer than a day \cite{brankovic2022identifying} were identified as influential predictors by the explainability algorithms. Except for the regression model (L1), all other methods included two predictors suggested by experts. One of the reasons for the unexpected result with L1 on the  RA30 outcome metric could be the particularity of the paediatric cohort. Some features recognised as relevant for these models, however, might be partially or completely non-representative for supporting clinical decision-making. For example, a patient’s admission source or elective status may not be as clinically relevant as recent pathology results. It was also observed that there was discordance between the features identified for both models across the explainability techniques employed. On average the agreement between each pair of methods was moderate across 5 independent runs. For the VS dataset, there was a degree of interdependence between the vital signs, hence clusters and associations of features that are not identical can indicate a common pattern or route of deterioration. \\

The potential actionability of the explanations relies on their capability to inform clinical workflow while being correct, parsimonious and timely.
In the context of predicting patient deterioration and risk of readmission, explanations obtained at the patient level were recognised by clinical collaborators as actionable, i.e. can guide the choice of intervention and can help busy clinicians prioritise their efforts while evaluating patients \cite{brankovic2022explainable}. However, while the studies have assessed model performance, there is no evidence supporting explanations reliability. To that end, one of the major contributions of this paper relates to the examination of the correctness of the generated explanations against the true causes recorded by the data collection system deployed in the study hospital (i.e. benchmarking), albeit only on one dataset (VS). On one hand, overall agreement with ground truth varied from moderate (48\% for XGB Shap) to strong (69\% and 73\% for SNN with shap and DTD explainers, respectively). On the other hand, if we exclude samples where the length of stay was identified as the most contributing feature, for XGB that is 40\% correctly predicted cases, the results on the remaining dataset were remarkably different. Matching on these samples, which listed vital signs as the main contributor, jumped to 75-80\% across the methods. This high level of  concordance is clinically relevant. As such the explanations are potentially reliable and actionable. Additionally, agreement on the top contributor across the methods for VS dataset is high (on average 84\%) including its sign agreement (99\%) which additionally contributes to building trust. Benchmarking explanations obtained for patient representation was impossible as there was no ground truth. Still, identified top predictors benefit from being already recognized as informative in predicting patient representation. Though, the agreement on the most contributing factor across the methods was 55\% which may affect clinicians' trust adversely. When considering timeliness, both algorithms can be leveraged to provide real-time predictions and explanations that provide relevant complementary information that is well aligned with the current clinical workflows, allowing for early intervention and the prioritisation of clinical efforts for care planning. 
While the initial observation of inconsistency and misalignment with the true triggers might give the impression that XAI is not useful to clinicians and cannot guide interventions, it is important to consider other factors before drawing a final conclusion. If we view explanations as a means to gain deeper insights while acknowledging their imperfections \cite{schwartz2022factors}, they can still provide valuable information to clinicians, especially when minimal trust is established. In situations where a clinician knows little about the patient’s condition, receiving an explanation could be a helpful suggestion, even if it is correct only half of the time because it provides some additional information. However, if the clinician is already highly confident that a patient is deteriorating in a recognizable, clinical pattern, providing an alternative explanation would require consideration in the same way alternative diagnoses or treatments would be considered.\\

% Consistency 
% -Sensitivity on underlying design variations, Sensitivity on deployed model, Agreement on the direction of the contribution
Findings in terms of consistency also varied. The Shap method demonstrated insensitivity to underlying design variations and obtained a moderate level of consistency with respect to the deployed model for the RA30 dataset and good for the VS dataset. Explanations obtained with the DTD method on top of DNN had the poorest consistency. From the clinical perspective, there are 3 sources of imperfection in the input features and the data available that could cause discrepancy and/or incorrectness of the results \cite{brankovic2023medinfo}: (i) Incomplete information to exactly “resolve” the question: missing features which, if available, would explain more of the variation and causation of the outputs. (ii) Dependence between factors. Several or most of the inputs have interrelationships with other input features and observations, e.g. in vital signs, pulse and blood pressure or respiratory rate, oxygen flow, and oxygen saturation. While these are each “independently related to the outcome” in the model, they are not independent of each other. This means that different groups of features could have the same implications for clinicians and hence the decision-making. Aas et al. \cite{aas2021explaining} showed  Shap \cite{lundberg2020local} may lead to incorrect explanations when features are highly correlated. (iii) Errors, missing data that could be observed or contradictory information. From the modelling perspective, the main reasons for discrepancy can be attributed to the optimisation objective which is simply minimisation of the prediction error. Consequently (i) Causal and/or clinically relevant associations might not be discovered and hence post hoc not explainable. (ii) Since different ML methods operate on different principles, different features may be identified as the most relevant.\\

The main limitation of the current study relates to benchmarking using explanations with only one dataset. For a more general conclusion, it is needed to validate the XAI on more datasets that contain the identification of the cause of future events such as the one we used in this study. This study considered only popular and readily available state-of-the-art XAI methods. However, exploring the consistency and concordance with recent methods that account for causality or multicollinearity e.g. [26, 32] and comparing the results to the well-established methods would be valuable  in future studies. The clinical implications of the results have been discussed with one clinical specialist. Findings from this study will be used to conduct an in-depth qualitative study to understand the interaction with clinicians’ judgment of the clinicians' trust and perception of the XAI-based support tools.

\section{Conclusion}
In this work, the trustworthiness and usefulness of explanations generated for predictive models in healthcare were examined. We evaluated  explanations against the metrics suggested \cite{tonekaboni2019clinicians} which could help to build trust in clinical settings. As part of it, we report benchmarking results, to the best of our knowledge first in this context.
This paper is suggesting that if 1) sufficient disparate ML methods agree on influential relationships, 2) if observable and input factors can be modified and they change model output, and 3) if the adjusted ML model outcomes concur and agree with the real-world results, then we are one step closer to trustworthy explanations. Unless these conditions are met, no explanation method should be considered trustworthy and or used to guide the choice of clinical intervention. However, they might still be useful in helping clinicians in cases when little or no information about patients is available or  to identify “at risk” patients who look deceptively stable.

\backmatter

\bmhead{Supplementary information}
This paper is accompanied by a supplementary file. 
% If your article has accompanying supplementary file/s please state so here. 

% Authors reporting data from electrophoretic gels and blots should supply the full unprocessed scans for key as part of their Supplementary information. This may be requested by the editorial team/s if it is missing.

% Please refer to Journal-level guidance for any specific requirements.

% \bmhead{Acknowledgments}

% Acknowledgments are not compulsory. Where included they should be brief. Grant or contribution numbers may be acknowledged.

% Please refer to Journal-level guidance for any specific requirements.

\section*{Declarations}

% Some journals require declarations to be submitted in a standardised format. Please check the Instructions for Authors of the journal to which you are submitting to see if you need to complete this section. If yes, your manuscript must contain the following sections under the heading `Declarations':

\bmhead{Competing interests} 
The authors have no competing interests to disclose. 

\bmhead{Availability of data and materials}
The datasets analysed in the current study are not publicly available. Due to reasonable privacy and security concerns and requirements imposed by the ethics approval process, they are not redistributable to researchers other than those engaged in the ethics committee-approved research protocol. Correspondence and requests for data should be addressed to A.B. (email: aida.brankovic@csiro.au).  

\bmhead{Code availability}
In this work we used following open-source libraries to conduct our experiments: For development of machine learning model we used sklearn (https://scikit-learn.org/) library developed for Python programming language. Shaply values used for explanations of the model were calculated with SHAP package (https://github.com/slundberg/shap). Adjusted to numerical inputs DTD explainer implemented in iNNvestigate (https://github.com/albermax/innvestigate) was used to analyse NN model. The analysis, training and tests were performed with custom code written in Python 3.6. 
All procedures and implementations are in Methods and supplement and allow identical replication. Any inquiry regarding the technical details of the specific models can be made by relevant parties to the corresponding author (email: aida.brankovic@csiro.au)

\bmhead{Authors' contributions}
A.B. initiated the project, provided project conception, and performed the analysis. D.C. was the clinical lead and provided clinical guidance in interpreting the results and discussing their implications. J.R. contributed to the literature review and the creation of a Supplementary information file. S.K. contributed results discussion. W.H. created a pipeline for the NN model and DTD explainer. A.B made the first article draft. All authors contributed to revising the first article draft and approving the final version of the manuscript.

\bibliography{sn-bibliography}% common bib file
%% if required, the content of .bbl file can be included here once bbl is generated
%%\input sn-article.bbl

\end{document}